\documentclass[conference]{IEEEtran} 
\IEEEoverridecommandlockouts 
\usepackage{verbatim}

\usepackage{amssymb}
\usepackage{amsmath}

\usepackage{multirow}

\usepackage{xskak}
\usepackage{chessboard}

\setchessboard{
smallboard,
showmover=false}

\newif\ifdebug

\newcommand{\gdl}[1]{\texttt{#1}}

\usepackage[utf8]{inputenc}
\usepackage[T1]{fontenc}

\ifCLASSINFOpdf
\else
\fi
\begin{document}
%
\title{Procedural Content Generation for GDL Descriptions of Simplified Boardgames}

%
%


\author{\IEEEauthorblockN{Jakub Kowalski}
\IEEEauthorblockA{Institute of Computer Science\\
University of Wroc{\l}aw, Poland\\
Email: jakub.kowalski@cs.uni.wroc.pl}
\and
\IEEEauthorblockN{Marek Szyku{\l}a}
\IEEEauthorblockA{Institute of Computer Science\\
University of Wroc{\l}aw, Poland\\
Email: marek.szykula@cs.uni.wroc.pl }
}

\maketitle

\begin{abstract}
We present initial research towards procedural generation of Simplified Boardgames and translating them into an efficient GDL code. This is a step towards establishing Simplified Boardgames as a comparison class for General Game Playing agents. To generate playable, human readable, and balanced chess-like games we use an adaptive evolutionary algorithm with the fitness function based on simulated playouts. In future, we plan to use the proposed method to diversify and extend the set of GGP tournament games by those with fully automatically generated rules.
\end{abstract}




%
\IEEEpeerreviewmaketitle


\section{Introduction} 

The increasing interest in algorithms which can play a variety of games, such as in the General Game Playing (GGP) \cite{Genesereth05,Love08}, or in the very new domain of General Video Game Playing (GVGP) \cite{Perez15}, should also be followed by  attempts to apply general procedural content generation for these domains. So far, automatic generation of game rules have been done for several classes of games, including, among others, commercially successful LUDI boardgames system \cite{Browne10}, Pell's Chess-Like Games \cite{Pell92}, turn-based strategy games \cite{Mahlmann11}, and games described by the Card Game Description Language \cite{Font13}.

GGP is a well establish area of research, with many contributions in different subdomains, and the level of agents attending the annual International General Game Playing Competition (IGGPC) is constantly increasing.
However, there is still no specified goal to be achieved by the GGP players, and the domain lacks an external comparison class. Our work intends to follow the idea from \cite{Kowalski15} to establish Bj\"{o}rnsson's Simplified Boardgames \cite{Bjornsson12} class (which is less general than GDL yet still interesting and challenging on its own) as a standard test class for GGP players.

As an important step in that direction, we propose a system for generating fair, chess-like boardgames and translating them into efficient GDL code. To ensure that generated boardgames fulfill desired criteria, we combine an evolutionary process of rules creation with simulation based fitness calculation. To properly rewrite generated boardgame into GDL, we use observations of minimizing description complexity for a game of chess described in \cite{Landau14}. 

We are convinced, that a method of procedural generation of GDL encoded Simplified Boardgames will spread automatically created boardgames as test games for evaluating level of GGP players.
Moreover, it could be even useful in the case of IGGPC, given that many games from that competition set are some kind of boardgames.


\section{Preliminaries} 

In this section we briefly introduce both game classes.

\subsection{General Game Playing and Game Description Language}

General Game Playing, in its modern form introduced in 2005 at Stanford University \cite{Genesereth05}, focuses on developing programs which can play, without human intervention, any game described by the Game Description Language. We assume that the reader is familiar with basics of this domain. For a detailed, up-to-date survey of the discipline we refer to~\cite{Genesereth14}.


Game Description Language (GDL) \cite{Genesereth05}, \cite{Love08}, is a strictly declarative language using logic programming-like syntax based on the Knowledge Interchange Format. Every game description contains declarations of player roles, initial game state, legal moves, state transition function, terminating conditions, and the goal function.

GDL has enough power to describe any turn-based, finite, deterministic, $n$-player game with full information and possibly simultaneous moves. What is important from the agent's point of view -- GDL does not provide any predefined functions: neither arithmetic expressions nor game-domain specific structures like board or card deck. Every function and declaration must be defined explicitly from scratch, and thereafter its semantic is given only by the game rules.

\subsection{Simplified Boardgames} 

The class of \emph{Simplified Boardgames} introduced by Bj\"{o}rnsson in \cite{Bjornsson12}, and slightly extended in \cite{Kowalski15}, describes rectangular board, turn-based, 2-player, zero-sum games with pieces movements being a subset of a regular language. It can describe many of the Fairy Chess variants, including the asymmetry and moves that capture own pieces. Possible winning conditions are reaching a fixed set of squares by a certain piece or capturing a fixed number of other player's pieces of certain type. However, there are some important limitations, mostly emerging from the regularity of the description, which makes moves impartial to the absolute location of the piece and to move history, so actions such as castling, en-passant, or promotions are not permitted.

A set of legal moves rules for each piece is the set of words described by a regular expression over an alphabet $\Sigma$ containing triplets $(\Delta x, \Delta y, on)$, where $\Delta x$, $\Delta y$ are relative column/row distances and $on \in\{e, p, w\}$ describes the content of the destination square: $e$ indicates an empty square, $p$ a square occupied by an opponent's piece, and $w$ a square occupied by an own piece. For example, capturing moves of rook are described by $(0,1,e)^*(0,1,p)+(0,-1,e)^*(0,-1,p)+(1,0,e)^*(1,0,p)+(-1,0,e)^*(-1,0,p)$. Knight's jump moves are for example $(1,2,e)$, $(-2, -1, p)$, etc. Given a current board situation, a move is legal if and only if it matches the content of the board. So if, for example, three consecutive squares in front of a white rook are empty, and there is a black pawn on the next one, the move $(0,1,e)^3(0,1,p)$ is legal. The rook can be moved by the vector $(0,4)$ and then the pawn on its final square will be captured.

\section{The Evolutionary System} 

We decided to focus our research on chess-like games to simplify the representation and increase the chance of well balanced instances. By chess-like, in this case we mean: initial position close to symmetrical with the line of weak pieces on the front of each army, fairy chess-like movements, and winning achieved by moving a piece to the opponent's backrank or taking all pieces of a given type.

The chromosome structure consists of three separate representations: for board, piece rules, and winning conditions. The board is a rectangular table containing in each square information about the piece's type and owner, or an empty square. Pieces are represented by a map from pieces symbols to its move rules. Winning conditions are encoded as a list of triples containing piece type, piece side, and the type of winning condition (capture or movement).

\subsection{Generating initial population}

At the beginning of the process, we generate building blocks for pieces movements. Based on some probability matrices we generate short words from $\Sigma^*$ and optionally assign them modifiers which extend a single word to a set of words. A modifier can mirror the move to be horizontally or vertically symmetrical, rotate the move, or add the Kleene star to any letter of the word.

We have divided types of pieces into three classes varying on mobility (inspired by the chess we have weak, light, and strong pieces). Using the building blocks we randomly build up sets of possible moves for every class. These sets are generated once for every evolutionary run.

The initial board position is randomized by a number of parameters. At first, a quarter of the board is generated with the probability of appearance of strong piece growing for the rows more distant from the opponent. Then, the generated quarter is mirrored (with some probability of change) to represent all pieces of the white player. After that, the white pieces are mirrored to the black ones, again with some probability of randomly regenerating a piece (with the constraint that it must be from the same class). 

Generating rules for pieces is relatively simple. For every piece type (e.g.\ pawn, rook, griffon) we just randomly choose its moves from the precomputed set for the corresponding class. Winning conditions are obtained by drawing piece type and capture/movement condition type. Possibilities of multiple winning conditions and asymmetry are provided by additional parameters.

\subsection{Fitness Function}

Before presenting the details of our evolutionary operators, we have to describe the method of evaluating each individual. As the main goal of evolution is very complex, and it cannot be easily expressed with a single function (which is a common problem with content generation for games), we use simulations to interpolate the features of the created game. 

The question is what are the features we want to acquire. Obviously, we want the game to be playable and as balanced as possible. Because generated games are chess-like, a large number of draws between players of comparable strength is permissible; however as we want the game to be strategic, we expect that better players should not have difficulty beating weaker ones (such games should also be much shorter). We also expect the games to be understandable and playable by humans, which requires that the rules are not too complex. Since we want them also to be playable by computers, the branching factor should also be restricted.

We apply two types of simulations (playouts): between two min-max algorithms, and between a min-max player against a random player (symmetrical for both roles). A min-max agent used for the comparison is the RBgPlayer, described in details in~\cite{Kowalski15}. We perform a fixed number of such simulations. From the results, the following game features are calculated:

\begin{itemize}
\item $P\in\{0, 1\}$ -- \emph{Playability}, which is 0 if the game is found unplayable, i.e.\ some simulation did not finish within the given timelimit, and 1 otherwise.
\item $L_M\in[0, 1]$ -- \emph{Min-max playouts length} which, for $\ell_{mm}$ being the mean length of playouts between two min-max players divided by the game turnlimit, is equal to $2\ell_{mm}$ for $l<\frac{1}{2}$, and 1 otherwise.
\item $L_D\in[0, 1]$ -- \emph{Playouts length difference} is calculated as follows.
Let $z = \ell_{mm} - \ell_{rm}$, where $\ell_{rm}$ is the mean length of playouts between the min-max player and the random player, divided by the game turnlimit.
The value of the feature is 1 if $z>\frac{1}{4}$, 0 if $z<0$ and $4z$ otherwise.
\item $W\in[-1,1]$ -- \emph{Width} is a normal distribution probability density rescaled to peak at 1, i.e.
\begin{equation} \label{eq:distribution}
W=e^{-\frac{(x-\mu)^2}{2\sigma^2}} ,
\end{equation}
where $\mu$ is equal to the given expected branching factor of the game, $\sigma$ is equal to $\frac{\mu}{4}$, and $x$ is the mean branching factor obtained from every state visited during all simulations.
\item $B_M\in[-1, 1]$ -- \emph{Min-max balance} is the balance between both players in min-max playouts calculated as $\mathit{white\_wins}-\mathit{black\_wins}$ divided by the number of playouts.
\item $B_R\in[-1, 1]$ -- \emph{Random balance} is the balance in the case of min-max vs.\ random games, i.e.\ $\mathit{white\_vs\_rand\_wins}-\mathit{black\_vs\_rand\_wins}$ divided by the number of min-max vs.\ random playouts.
\item $S\in[0, 1]$ -- \emph{Strategy} describes chances of winning against the weak opponent, and is calculated as the number of wins of min-max vs.\ random player divided by the number of such playouts.
\item $C\in[0, 1]$ -- \emph{Complexity} is calculated based on the size of the games rules compared to some desired size $s$ (which is based on the size of chess pieces descriptions and depends on the number and class of pieces). The feature is equal to the absolute value of~(\ref{eq:distribution}), where $\mu$ is the desired game size, $\sigma$ is equal to $\frac{\mu}{4}$, and $x$ is the actual game size.
\item $R\in[0, 1)$ -- \emph{Reducibility} means that not all pieces are really important for the game. Assume that there are $k$ pieces, and $u_i$ is the \emph{usefulness} measure of $i$-th piece, calculated as the number of moves performed with this piece by min-max players, divided by the number of all moves made by min-max players. Then, this feature is equal to $\sum_{i=1}^k (\frac{1}{k}-u_i)^2$
\end{itemize}

Based on these features, the game's fitness is calculated using the following formula:
{\small\begin{equation*}
\frac{P}{15}\left(L_M + L_D + 2|W| + 3S + 2C + (6-3|B_M|-|B_R|-2R)\right).
\end{equation*}}

\subsection{Evolving}

We start with the initial population of size $n$. Every individual is evaluated, and if it is found unplayable it is removed from the population. The remaining games are ordered by $B_M+B_R$ measure. Parents are chosen by pairing games from the ends to the center, i.e.\ the first with the last, the second with the last but one, etc. Crossover creates two offsprings in the following way. We use two point crossover on a board columns, and a uniform crossover on pieces rules and winning conditions (the conditions may have different length for both parents).

We also mutate every individual from the population in an adaptive GA fashion. Let $s$ be the sum of the board's dimensions. We mutate pieces rules and winning conditions with probability $\frac{1}{s}$, and the initial position otherwise. The initial position is mutated by randomly choosing a square and its new content (empty or a piece of random type). If the square is on the lower half of the board, the piece is white, otherwise it is black. In the case of mutating pieces rules we have three options depending on $R$, $C$, and some constant weight. We can draw new rules for a piece with the lowest usability, or choose it using roulette wheel based on the complexity of the piece's rules. The last option is to swap the rules of two pieces of the same class. Mutation of terminal conditions is guided by balance values. With some constant weight, both sides have added or removed a winning condition. With a weight dependent on $B_M$ and $B_R$, only one side's conditions are modified, to possibly improve game's balance.

The mutants and offsprings are evaluated and removed if unplayable. For the next generation we take best $n$ individuals from the joint set of parents, offsprings, and mutants. If there is only $k<n$ playable games, $n-k$ best individuals are duplicated.


\section{Experiment result}

We performed several evolutionary runs with different board sizes (from $6\times6$ to $8\times8$), numbers of pieces per class, population sizes, and numbers of playouts. In the case with $6\times6$ board and population of size 50, we were able to evolve an entire population with the fitness above 0.98. With a smaller population and larger boards, the evolution usually stuck at a local maximum around 0.85--0.88. The two features which we found the most difficult to optimize were complexity and reducibility. 


We present one of the evolved games, appeared in the 22-nd round of evolution, with fitness 0.9899. We have found this game interesting because of non-trivial asymmetry, simple and interesting pieces, and good reducibility. The game was also rated as fully balanced, with all playouts lost by the random player, and 20\% wins for both white and black sides in the case of min-max vs.\ min-max playouts.

The game uses $6\times6$ board with two weak pieces (\textsymfigsymbol{P}\textsymfigsymbol{K}), two light pieces (\textsymfigsymbol{N}\textsymfigsymbol{B}), and one strong piece (\textsymfigsymbol{Q}). The turnlimit is 60 and there are no other winning conditions specified, which means that one has to capture all enemy pieces. (We have found this type of victory surprisingly common among best-fitted individuals.) The initial position looks as follows:

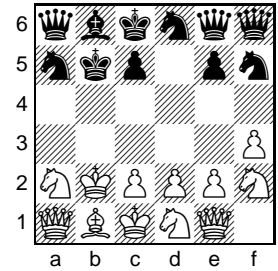
\begin{figure}[!h]
\centering
\chessboard[maxfield=f6,,setpieces={Qa1,Bb1,Kc1,Nd1,Qe1,Na2,Kb2,Pc2,Pd2,Pe2,Nf2,Pf3,na5,kb5,pc5,pe5,nf5,qa6,bb6,kc6,nd6,qe6,qf6}]
\end{figure}

The pieces with the biggest usefulness (0.28) are \textsymfigsymbol{N} and \textsymfigsymbol{Q}, while the least useful pieces are \textsymfigsymbol{K} (0.12) and \textsymfigsymbol{B} (0.13). The pieces rules are as follows:

\begin{itemize}
 \item[\textsymfigsymbol{P}] Moves and captures 1 cell diagonally ahead.
 \item[\textsymfigsymbol{K}] Moves and captures 1 or 2 squares straight ahead, jumping the intervening square in the case of moving 2 squares.
 \item[\textsymfigsymbol{N}] Moves and captures 1 cell diagonally, and it can jump orthogonally forward by 2 squares if the destination is empty.
 \item[\textsymfigsymbol{B}] Moves like the orthodox bishop, and it can move and capture 1 square straight ahead.
 \item[\textsymfigsymbol{Q}] Slides any number orthogonally if all pieces on the way are the opponent's, and it can capture and self-capture 1 case diagonally ahead, and move and self-capture 1 square straight-ahead, and 1 square straight-back. 
\end{itemize}

\newpage 
\section{Translation from Simplified Boardgames to GDL}

To make the class of Simplified Boardgames useful to GGP agents, it is essential to provide a translation to GDL (see~\cite{Kowalski14} for a recent translation from the Card Game Description Language). This would not only make it possible to compare GGP agents with more specialized programs (cf.~\cite{Kowalski15}), but also provide a source of automatically generated games for tests and competitions. This is of certain importance, since so far almost all games in GGP are human-prepared.

We plan to design and implement a translator for Simplified Boardgames. Our aim is to generate possibly efficient GDL code, making translated games more accessible to GDL agents. To do so, we plan to make use of the structure of the language of Simplified Boardgames and various tricks commonly used in encoding board games in GDL~\cite{Landau14}.

The translation can be done in a natural way. Here let us just consider computation of legal moves, which is the most challenging part. First, we can create a predicate storing the content on each occupied square: \gdl{\textbf{true}(board($x$,$y$,$\mathit{piece}$,$\mathit{owner}$))}. For each regular expression $(\Delta x_1,\Delta y_1,on_1)\dots(\Delta x_k,\Delta y_k,on_k)$ defining legal moves of $\mathit{piece}$, we create the following corresponding rule:

\noindent\gdl{\textbf{legal}(?owner,move(?x0,?y0,?xk,?yk)) $\Leftarrow$\\
\indent \textbf{true}(board(?x0,?y0,$\mathit{piece}$,?owner)) $\wedge$\\
\indent sum(x0,$\Delta x_1$,?x1) $\wedge$\\
\indent sum(y0,$\Delta y_1$,?y1) $\wedge$\\
\indent \textbf{true}(board(?x1,?y1,?dummy1,$on_1$))$\wedge$\\
\indent \dots $\wedge$\\
\indent sum(x$(k-1)$,$\Delta x_k$,?x$k$) $\wedge$\\
\indent sum(y$(k-1)$,$\Delta y_k$,?y$k$) $\wedge$\\
\indent \textbf{true}(board(?x$k$,?y$k$,?dummy$k$,$on_k$))}.

Here \gdl{sum} is a constant predicate commonly used to simulate arithmetic in GDL.
Note also that the stars and powers in regular expressions can be expanded and translated in a similar way.

This standard approach would be, however, not very effective. It involves large domains (note the number of possible facts in the domain of the board predicate) and the rules are expensive to process. There are several ways to optimize that. We can observe that, in most cases, regular expressions allow the sharing of some computations, e.g. for $(1,2,e)(2,0,e)+(1,2,e)(2,3,e)$ we can can use an additional temporary predicate storing intermediate moves by $(1,2,e)$. The same concept becomes more powerful when we deal with stars; these translate naturally to cyclic rules. It may be also worthwhile to consider specific kinds of moves that are shared among various pieces (e.g. rook and queen in chess). These considerations would lead to an algorithm that would compute possibly the simplest set of rules covering all moves.

An alternative approach is to define identifiers for each piece and store their positions in separate predicates (e.g.\ \gdl{rook\_white\_\textit{id}($x$,$y$)}), rather than to store the whole board in a single one.
Assuming that the GGP agent is able to infer that there can be at most one fact for given parameters (one piece on one square), it may result in much smaller domains, when there are only a few of pieces relatively to the board size. In the former approach we have at least $(nm)^{2\#t}$ possible facts in the board predicate, where $\#t$ is the number of types of pieces, and $n m$ is the size of the board. The latter approach gives us $\#p^{n m}$, where $\#p$ is the number of pieces on the board at the beginning. The translator can decide which approach would be more efficient, basing on an analysis of the complexity of the game.

\section{Conclusions} 

This paper presents an evolutionary system that automatically generates Simplified Boardgames. Generated games are evaluated using simulations with min-max and random reference players. Games are rated using several features measuring balance, strategy influence, pieces usefulness, game tree size, and rules complexity. An example of the evolved, balanced, chess-like, and human readable game is provided.

We plan to further improve the system and make it more general. In particular, it should be able to generate fully asymmetric non chess-like games. The issue of much importance is the fitness function, and in future versions we want it to be based on \emph{learnablity}, defined as a potential of improving the level of play related to the number of games played. By adding to the evolver an algorithm rewriting boardgames into GDL, we are going to extend the experiments from~\cite{Kowalski15} and test GGP agents on automatically generated games.

\bibliographystyle{IEEEtran}
\bibliography{bibliography}
%



\end{document}